\documentclass[letterpaper]{article} 
\usepackage{aaai20}  
\usepackage{times}  
\usepackage{helvet} 
\usepackage{courier}  
\usepackage[hyphens]{url}  
\usepackage{graphicx} 
\usepackage{graphicx}  
\usepackage{color}
\usepackage{amsmath}
\usepackage{amsfonts}
\usepackage{algorithm}
\usepackage[noend]{algpseudocode}
\usepackage{enumitem}
\usepackage{booktabs}
\usepackage{multirow}
\usepackage{colortbl}
\usepackage{soul}

\newcommand{\unk}{{\small [UNK] }}
\newcommand{\rowc}{\rowcolor[gray]{0.9}}
\newcommand{\citet}[1]{\citeauthor{#1}~\shortcite{#1}}

\renewcommand{\paragraph}[1]{\textbf{#1}}

\newcommand{\mysout}[1]{\setlength{\fboxsep}{0pt}\colorbox{black}{\strut #1}}

\title{Assessing the Benchmarking Capacity of \\ Machine Reading Comprehension Datasets}

\author{
  Saku Sugawara$^*$ \\
  University of Tokyo \\
  sakus@is.s.u-tokyo.ac.jp
  \And
  Pontus Stenetorp \\
  University College London \\
  p.stenetorp@cs.ucl.ac.uk
  \AND
  Kentaro Inui \\
  Tohoku University and RIKEN Center for AIP \\
  inui@ecei.tohoku.ac.jp
  \And
  Akiko Aizawa \\
  National Institute of Informatics \\
  aizawa@nii.ac.jp
}

\begin{document}

\maketitle

\begin{abstract}
  Existing analysis work in machine reading comprehension (MRC) is largely concerned with evaluating the capabilities of systems.
  However, the capabilities of datasets are not assessed for benchmarking language understanding precisely.
  We propose a semi-automated, ablation-based methodology for this challenge;
  By checking whether questions can be solved even after removing features associated with a skill requisite for language understanding, we evaluate to what degree the questions do \textit{not} require the skill.
  Experiments on 10 datasets (e.g., CoQA, SQuAD v2.0, and RACE) with a strong baseline model show that, for example, the relative scores of the baseline model provided with \textit{content words only} and with \textit{shuffled sentence words} in the context are on average 89.2\% and 78.5\% of the original scores, respectively.
  These results suggest that most of the questions already answered correctly by the model do not necessarily require grammatical and complex reasoning.
  For precise benchmarking, MRC datasets will need to take extra care in their design to ensure that questions can correctly evaluate the intended skills.
\end{abstract}

\section{Introduction}
\label{sec:intro}

\begin{figure}[tb]
  \newcommand{\highlight}[1]{\textcolor{blue}{\textbf{#1}}}
  \newcommand{\highlightx}[1]{\textcolor{red}{#1}}
  \newcommand{\miniindent}{0.98\linewidth}
  \newenvironment{miniright}{
    \vspace{-0.2em}\begin{flushright}\begin{minipage}[t]{\miniindent}
  }{
    \end{minipage}\end{flushright}\vspace{-0.2em}
  }
  \newcommand{\empA}[1]{\underline{#1}}
  \newcommand{\empB}[1]{\textit{\textbf{#1}}}
  \centering \footnotesize
  \fbox{%
    \parbox{0.995\linewidth}{%
      \textbf{Original context} 
      \begin{miniright}
        Immediately behind the basilica is the Grotto, a Marian place of prayer and reflection. It is a replica of the grotto at Lourdes, France where the Virgin Mary reputedly appeared \empB{to} \empA{Saint Bernadette Soubirous} \empB{in 1858}. At the end of the main drive (and in a direct line that connects through 3 statues and the Gold Dome), is a simple, modern stone statue of Mary.
      \end{miniright}

      \textbf{Anonymized context}
      \begin{miniright}
        @adv1 @prep5 @other0 @noun17 @verb2 @other0 @noun20 @punct0 @other1 @adj3 @noun21 @prep1 @noun22 @other2 @noun23 @period0 @other3 @verb2 @other1 @noun24 @prep1 @other0 @noun20 @prep6 @noun25 @punct0 @noun26 @wh0 @other0 @noun7 @noun8 @adv3 @verb4 \empB{@prep4} \empA{@noun27 @noun28 @noun29} \empB{@prep2 @num0} @period0 @prep6 @other0 @noun30 @prep1 @other0 @adj4 @noun31 @punct3 @other2 @prep2 @other1 @adj5 @noun32 @wh1 @verb5 @prep7 @num1 @noun6 @other2 @other0 @noun4 @noun5 @punct4 @punct0 @verb2 @other1 @adj6 @punct0 @adj7 @noun33 @noun6 @prep1 @noun8 @period0
      \end{miniright}

      \textbf{Question}
      \begin{miniright}
        \empB{To} whom did the Virgin Mary allegedly appear \empB{in 1858} in Lourdes France?
      \end{miniright}

      \textbf{Anonymized question}
      \begin{miniright}
        \empB{@prep4} @wh2 @verb6 @other0 @noun7 @noun8 @adv4 @verb4 \empB{@prep2 @num0} @prep2 @noun25 @noun26 @period1
      \end{miniright}
      \textbf{Baseline model's prediction before / after anonymization}
      \begin{miniright}
        \empA{Saint Bernadette Soubirous} / \empA{noun27 @noun28 @noun29}
      \end{miniright}
    }
  }
  \caption{
    Example of an ablation test that anonymizes context and question words, applied to a question from SQuAD v1.1 \cite{rajpurkar2016squad} with the correct answer in underscored.
    We found that the baseline model can achieve 61.2\% F1 on SQuAD v1.1 even after the anonymization.
  }
  \label{fig:intro}
\end{figure}

Machine reading comprehension (MRC) is a testbed for evaluating natural language understanding (NLU), by letting machines answer questions about given texts \cite{hirschman1999deep}. 
Although MRC could be the most suitable task for evaluating NLU \cite{chen2018neural} and the performance of systems is comparable to humans on some existing datasets \citet{devlin2019bert}, it has been found that the quality of existing datasets might be insufficient for requiring precise understanding \cite{jia2017adversarial}. 
Whereas these analyses are useful to investigate the performance of \textit{systems}, however, it is still necessary to verify the fine-grained capabilities of \textit{datasets} for benchmarking NLU.


In the design of MRC datasets, it is implicitly assumed that questions test a cognitive process of language understanding \cite{sutcliffe2013QA4MRE}.
As various aspects of such a process, we can use \textit{requisite skills} for answering questions such as coreference resolution and commonsense reasoning \cite{sugawara2017evaluation}. 
Considering skills as metrics would be useful for analyzing datasets.
However, for most datasets, the skills required to answer existing questions are not identified, or significant human annotation is needed. 
In this study, we propose a semi-automated, ablation-based methodology to analyze the capabilities of MRC datasets to benchmark NLU.
Our motivation is to investigate to what extent a dataset allows unintended solutions that do not need requisite skills.
This leads to the following intuition: if a question is correctly answered (or \textit{solvable}) even after removing features associated with a given skill, the question does not require the skill.
We show an example of our ablation method in Figure \ref{fig:intro}.
Suppose we wish to analyze a dataset's capacity to evaluate understanding of texts beyond the information of part-of-speech (POS) tags.
To this end, we replace context and question words with POS tags and ID numbers.
If a model can still correctly answer this modified question, the question does not necessarily require deep understanding of texts but matching word patterns only.
Questions of this kind might be insufficient for developing a model that understands texts deeply as they may reduce models to recognizing superficial word overlaps.

Our methodology uses a set of requisite skills and corresponding ablation methods.
Inspired by the computational model of reading comprehension \cite{kintsch1988role}, we exemplify 12 skills on two classes: reading and reasoning (Section~\ref{sec:skills}).
%
Then, we present a large-scale analysis over 10 existing datasets using a strong baseline model 
(Section~\ref{sec:experiment}). 
%
%
In Section~\ref{sec:human}, 
we perform a complementary inspection of questions with our ablation methods in terms of the solvability of questions and the reconstructability of ablated features.
Finally we discuss, in Section~\ref{sec:discussion}, two requirements for developing MRC to benchmark NLU: the control of question solvability and the comprehensiveness of requisite skills.

Our contributions are as follows:

\begin{itemize}
\item We propose a semi-automated methodology to analyze the benchmarking capacity of MRC datasets in terms of requisite skills for answering questions. 
\item With an example set of 12 skills and corresponding input-ablation methods, we use our methodology and examine 10 existing datasets with two answering styles. 
\item Our analysis shows that the relative performance on questions with \textit{content words only}, \textit{shuffled sentence words}, and \textit{shuffled sentence order} averaged 89.2\%, 78.5\%, and 95.4\% of the original performance, indicating that the questions might be inadequate for evaluating grammatical and complex reasoning. 
\end{itemize}

These results suggest that most of the questions currently \textit{solved} in MRC may be insufficient for evaluating various skills.
A limitation of our method is that it can not draw conclusions regarding questions that remain \textit{unsolved}, and thus we need to assume a reasonable level of performance for existing models on the dataset to be analysed.
Given our findings, we posit that MRC datasets should be carefully designed, e.g., by filtering questions using methods such as the ones we propose, so that their questions correctly benchmark the intended NLU skills.

\section{Related Work}
\label{sec:related}

We briefly survey existing interpretation methods and skill-based analyses for NLU tasks.

\paragraph{Interpretation methods.}
A challenge with the MRC task is that we do not know the extent to which a successful model precisely understands natural language.
To analyze a model's behavior, existing studies mainly proposed modification of the input.
For example, \citet{jia2017adversarial} showed that the performance of existing models on SQuAD \cite{rajpurkar2016squad} significantly degrades when manually verified distracting sentences are added to the given context.
In addition, 
\citet{feng2018pathologies} demonstrated that MRC models do not necessarily change their predictions even when most question tokens are dropped.
Likewise, for the natural language inference task, \citet{gururangan2018annotation} proposed to hide the premise and to evaluate a model using only the hypothesis.
These kinds of analyses are helpful for detecting biases that are unintentionally included in datasets.
Nonetheless, to assure that a dataset can evaluate various aspects of NLU, more fine-grained detail is needed than what is allowed by inspection using existing methods.

\paragraph{Skills as units of interpretation.}
In the topic of interpretable machine learning, \citet{doshi2018considerations} defined the concept of \emph{cognitive chunks} as the basic units of explanation.
In the MRC task, we consider that \textit{requisite skills} to answer questions are appropriate as such units.
A skill-based analysis was conducted by \citet{boratko2018systematic}, who proposed classifications of knowledge and reasoning. 
Prior to this, \citet{sugawara2017evaluation} also defined a set of 13 requisite skills. 
However, there are two main issues with these approaches: (i) the human annotation does not necessarily reveal unintended biases that machines can make use of, 
and (ii) it requires costly annotation efforts.
Therefore, we posit that a machine-based analysis is needed and that it should be performed in an automated manner.

\section{Dataset Diagnosis by Input Ablation}
\label{sec:skills}

\begin{table*}[t]
  \normalfont \centering
  \newcommand{\marginbreak}{\\[0.15em]}
  \setlength{\tabcolsep}{6pt}
  \newcounter{rowcounter}
  \newcommand{\rownumber}{\stepcounter{rowcounter}\ifnum\value{rowcounter}<10 \hphantom{0}\fi\arabic{rowcounter}.}
    \begin{tabular}{c @{\hspace{0.5\tabcolsep}} c @{\hspace{0.5\tabcolsep}} p{17em} p{28.2em}} \toprule
    & & Comprehension skill $s_i$ & Ablation method $\sigma_i$ \\ 
    \midrule
    \parbox[t]{2mm}{\multirow{8}{*}{\rotatebox[origin=c]{90}{Reading-class}}} & \rownumber & Recognizing question words excluding interrogatives & Drop all words except interrogatives (\emph{wh-} words and \emph{how}) in a question. \marginbreak
    & \rownumber & Recognizing content words & Drop content words in the context. \marginbreak
    & \rownumber & Recognizing function words & Drop function words in the context. \marginbreak
    & \rownumber & Recognizing vocabulary & Anonymize context and question words with their part-of-speech tag. \marginbreak
    & \rownumber & Attending the whole context other than similar sentences &  Keep the sentences that are the most similar to the question in terms of unigram overlap and drop the other sentences. \marginbreak
    & \rownumber & Recognizing the word order & Randomly shuffle all words in the context. \\
    \midrule
    \parbox[t]{2mm}{\multirow{6}{*}{\rotatebox[origin=c]{90}{Reasoning-class~}}} & \rownumber & Grasping sentence-level compositionality & Randomly shuffle the words in all the sentences except the last token. \marginbreak 
    & \rownumber & Understanding of discourse relations & Randomly shuffle the order of the sentences in the context. \marginbreak
    & \rownumber & Performing basic arithmetic operations & Replace numerical expressions (CD tag) with random numbers. \marginbreak 
    & \rownumber & Explicit logical reasoning & Drop logical terms such as \emph{not}, \emph{every}, and \emph{if}. \marginbreak
    & \rownumber & Resolving pronoun coreferences & Drop personal and possessive pronouns (PRP and PRP\$ tags). \marginbreak 
    & \rownumber & Reasoning about explicit causality & Drop causal terms/clauses such as \emph{because} and \emph{therefore}. \\
    \bottomrule
  \end{tabular}
  \caption{
    Example set of requisite skills $\{s_i\}$ and corresponding ablation methods $\{\sigma_i\}$.
    $f$ is a model and $(x, y)$ is a pair consisting of an input instance and its gold-standard answer.
    We interpret that for $x$ s.t. $f(x) = y$, if $f(\sigma_i(x)) = y$, then $x$ is solvable without $s_i$.
  }
  \label{tbl:ablation}
\end{table*}

\subsection{Formulation}

Our methodology uses a set of requisite skills and corresponding ablation methods.
By checking the solvability of questions after applying the ablation methods, we can quantify to what degree the questions allow unintended solutions that do not require the requisite skills. 
Users can define an arbitrary set of skills to suit their purposes. 

We develop a method $\sigma_i$ that ablates features necessary for the corresponding skill $s_i$ in a set of requisite skills $S$.
For $(x, y) \in X \times Y$, whenever $f(x) = y$, if $f(\sigma_i(x)) = y$, we recognize that $x$ is solvable without $s_i$.
\noindent Here, $X$ is the input, $Y$ is the gold labels, $(x, y)$ is a pair consisting of an input instance and its gold-standard answer, and $f$ is a model.
When the performance gap between the original and the modified dataset is small, we can infer that most of the questions already solved are solvable without $s_i$.
On the other hand, if the gap is large, a sizable proportion of the solved questions may require $s_i$.

We note that we cannot draw general conclusions for instances given by conditions other than the abovementioned one.
Consider the case where $f(x) = y$ and $f(\sigma_i(x)) \neq y$, for example.
This only means that $f$ cannot solve $x$ without the features ablated by $\sigma_i$.
We cannot conclude that $x$ requires $s_i$ in \emph{every} model because there might exist a model that can solve $x$ without $s_i$.
However, if there is \emph{at least one} model $f$ that solves $x$ without $s_i$, this may indicate an unintended way to solve $x$ while ignoring $s_i$. 
Therefore our methodology only requires a single baseline model.
Users can choose an arbitrary model for their purposes.

\subsection{Example Set of Requisite Skills}

In this section, we exemplify a skill set that consists of 12 skills along with two classes; reading and reasoning (Table~\ref{tbl:ablation}).
In psychology, there is a tradition of theoretical research on human text comprehension.
The construction--integration model \cite{kintsch1988role} is one of the most acknowledged theories.
This model assumes that human text comprehension consists of two processes: (i) construction, in which a reader elaborates concepts and propositions in the text and (ii) integration, in which the reader associates the propositions to understand them consistently.
We associate this two-step process with our two classes. 

\textbf{Reading skills.}
This class deals with six skills of observing and recognizing word appearances, which are performed before reasoning.
In MRC, it has been shown that some existing questions can be solved by reading a limited number of words in the question and the context (e.g., by simply attending to context tokens that are similar to those of the questions \cite{sugawara2018what}).
Our goal of this class is, therefore, to ensure that the questions require the reading of the whole question and context uniformly.

\textbf{Reasoning skills.}
This class comprises six skills of relational reasoning among described entities and events such as pronoun coreference resolution and logical reasoning.
Although these skills are essential for sophisticated NLU, it is difficult to precisely determine whether these types of reasoning are genuinely required in answering a question. 
Therefore, in this class, we define reasoning-related skills that are performed using the \textit{explicit} information contained in the context (e.g., $s_9$ explicit logical reasoning and $s_{12}$ reasoning about explicit causality).

In the following, we highlight some of the defined skills.
Skill $s_1$ is inspired by \citet{feng2018pathologies} and \citet{sugawara2018what}.
Although their studies proposed dropping question tokens based on their model-based importance or the question length, we instead drop tokens other than interrogatives as interpretable features. 
Our vocabulary anonymization ($s_4$) is mainly inspired by \citet{hermann2015teaching} where they anonymized named entities to make their MRC task independent of prior knowledge.
Our shuffle-based methods ($s_6$ to $s_8$) are inspired by existing analyses for other tasks \cite{khandelwal2018sharp,nie2019analyzing,sankar2019neural}.
Among them, our purpose for $s_7$ is to analyze whether a question requires \emph{precise} reasoning performed over syntactic and grammatical aspects in each sentence. 
The remaining skills are described in Appendix A.
Although our proposed definitions can be extended, they are sufficient for the purpose of demonstrating and evaluating our approach.
In Section~\ref{sec:discussion}, we discuss further directions to develop purpose-oriented skill sets.

\section{Experiments and Further Analyses}
\label{sec:experiment}

\subsection{Experimental Settings}

\paragraph{Datasets.}
We use 10 datasets.
For answer extraction datasets in which a reader chooses a text span in a given context, we use (1) CoQA \cite{reddy2019coqa}, (2) DuoRC \cite{saha2018duorc}, (3) HotpotQA (distractor) \cite{yang2018hotpotqa}, (4) SQuAD v1.1 \cite{rajpurkar2016squad}, and (5) SQuAD v2.0 \cite{rajpurkar2018know}.
For multiple choice datasets in which a reader chooses a correct option from multiple options, we use (6) ARC (Challenge) \cite{clark2018think}, (7) MCTest \cite{richardson2013MCTest}, (8) MultiRC \cite{khashabi2018looking}, (9) RACE \cite{lai2017race}, and (10) SWAG \cite{zellers2018swag}.
For the main analysis, we applied our ablation methods to development sets.
We included SWAG because its formulation can be viewed as a multiple-choice MRC task and we would like to analyze the reasons for the high performance reported for the baseline model on this dataset \cite{devlin2019bert}.
For preprocessing the datasets, we use CoreNLP \cite{manning2014stanford}.
We specify further details in Appendix~B.

\paragraph{Models.}
As the baseline model, we used BERT-large \cite{devlin2019bert}.\footnote{
  Although our methodology only necessitates a single baseline model, note that we need to assume a reasonable level of performance as we mentioned in Section \ref{sec:intro}.
}
We fine-tuned it on the original training set of each dataset and evaluated it on a modified development set.
For $\sigma_4$ vocabulary anonymization, we train the model after the anonymization.
For ARC, MCTest, and MultiRC, we fine-tuned a model that had already been trained on RACE to see the performance gained by transfer learning \cite{sun2019improving}.
We report the hyperparameters of our models in Appendix~C.
Although we trained the baseline model on the original training set, it is assumed that the upper-bound performance can be achieved by a model trained on the modified training set.
Therefore, in Section \ref{sec:further}, we also see the extent to which the performance improves when the model is trained on the modified training set. 

\paragraph{Ablation methods.}
$\sigma_2$ and $\sigma_3$: we use a set of stopwords from NLTK \cite{loper2002nltk} as function words.
All other words are regarded as content words.
We do not drop punctuation.
When a token is dropped, it is replaced with an \unk token to preserve the correct answer span.
$\sigma_4$: we use the same ID for the same word in a single given context but different IDs for different contexts. For inflectional words, we anonymize them using their lemma. For example, \textit{are} would be replaced with \textit{@verb2} (= \textit{is}) if it appeared in Figure~\ref{fig:intro}. In addition, to retain the information of the POS tags, we append its POS tag after each inflectional anonymized word (e.g., \textit{is} is replaced with \textit{@verb\{ID\} [VBZ]}).
$\sigma_6$: because it is necessary to maintain the correct answer span in the answer extraction datasets, we split the context into segments that have the same length as the gold answer span and shuffle them.
$\sigma_7$: as with $\sigma_6$, we split each sentence into segments and shuffle them within each sentence.
For $\sigma_6$ to $\sigma_8$, we averaged the scores over five runs with different seeds and report their variances in Appendix~D.

\begin{table*}[th]
  \footnotesize
  \newcounter{rowcounter2}
  \setlength{\tabcolsep}{4pt}
  \def\arraystretch{1.1}
  \newcommand{\rc}[1]{\hspace{-0em}$_\text{{#1}}$}
  \newcommand{\rownumber}{\stepcounter{rowcounter2}\ifnum\value{rowcounter2}<10 \hphantom{0}\fi\arabic{rowcounter2}. }
  \newcommand{\linestack}[1]{\def\arraystretch{0.8}\begin{tabular}[c]{@{}c@{}} #1 \end{tabular}}
  \begin{tabular}{lcccccccccc|c} \toprule
    Ablation method \textbackslash ~Dataset & CoQA & DuoRC & \linestack{Hotpot- \\ QA} & \linestack{SQuAD \\ v1.1} & \linestack{SQuAD \\ v2.0} & ARC & MCTest & \linestack{Multi- \\ RC} & RACE & SWAG & \linestack{~Rel. \\ ~avg.} \\ \midrule
    \hphantom{00.} Answering style & \multicolumn{5}{c|}{answer extraction (F1)} & \multicolumn{5}{c}{multiple choice (accuracy)} \\
    \midrule
    \hphantom{00.}
    Original dataset    & 77.4\rc{\hphantom{-0}0.0} & 58.4\rc{\hphantom{-0}0.0} & 63.6\rc{\hphantom{-0}0.0} & 91.5\rc{\hphantom{-0}0.0} & 81.9\rc{\hphantom{-0}0.0} & 52.7\rc{\hphantom{-0}0.0} & 87.8\rc{\hphantom{-0}0.0} & 78.0\rc{\hphantom{-0}0.0} & 68.8\rc{\hphantom{-0}0.0} & 85.4\rc{\hphantom{-0}0.0} & \hphantom{-0}0.0  \\
    \midrule
        \rowc \rownumber Q interrogatives only      & 20.1\rc{-74.0} & 14.2\rc{-75.8} & 15.0\rc{-76.4} & 15.2\rc{-83.4} & 50.1\rc{-38.9} & 35.6\rc{-32.5} & 64.1\rc{-27.0} & 52.6\rc{-32.6} & 56.7\rc{-17.5} & 77.1\rc{\hphantom{0}-9.7} & -46.8  \\
         \rownumber Function words only      & 53.0\rc{-31.5} & \hphantom{0}5.8\rc{-90.1} & \hphantom{0}7.8\rc{-87.8} & 17.4\rc{-81.0} & 50.2\rc{-38.7} & 44.0\rc{-16.6} & 32.2\rc{-63.3} & 61.9\rc{-20.6} & 43.2\rc{-37.3} & 68.9\rc{-19.4} & -48.6  \\
        \rowc \rownumber Content words only             & 60.9\rc{-21.3} & 47.9\rc{-18.0} & 56.2\rc{-11.6} & 80.7\rc{-11.8} & 73.5\rc{-10.3} & 48.0\rc{\hphantom{0}-8.9} & 80.3\rc{\hphantom{0}-8.5} & 74.5\rc{\hphantom{0}-4.5} & 62.0\rc{\hphantom{0}-9.8} & 82.6\rc{\hphantom{0}-3.3} & -10.8  \\
        \rownumber Vocab. anonymization  &  39.0\rc{-49.6} & 18.6\rc{-68.2} & 16.8\rc{-73.6} & 61.2\rc{-33.1} & 59.4\rc{-27.0} & 29.2\rc{-44.6} & 25.3\rc{-71.2} & 57.2\rc{-26.7} & 26.1\rc{-62.1} & 25.5\rc{-70.1} & -52.6 \\
\rowc \rownumber Most sim. sent. only     & 32.6\rc{-57.9} & 35.8\rc{-38.7} & 16.9\rc{-73.4} & 68.5\rc{-25.1} & 72.8\rc{-11.2} & 43.6\rc{-17.2} & 50.3\rc{-42.7} & 67.9\rc{-12.9} & 52.1\rc{-24.3} & 85.4\rc{\hphantom{0}-0.1} & -30.4  \\
\rownumber Context words shuff.        & 29.8\rc{-61.5} & 25.4\rc{-56.6} & 23.6\rc{-62.9} & 35.9\rc{-60.7} & 52.4\rc{-36.1} & 47.4\rc{\hphantom{0}-9.9} & 47.2\rc{-46.3} & 64.3\rc{-17.6} & 51.7\rc{-24.9} & 78.6\rc{\hphantom{0}-8.0} & -38.4  \\
\rowc  \rownumber Sentence words shuff.    & 53.0\rc{-31.6} & 35.9\rc{-38.6} & 43.1\rc{-32.2} & 62.1\rc{-32.1} & 64.4\rc{-21.4} & 46.4\rc{-11.8} & 70.6\rc{-19.6} & 71.4\rc{\hphantom{0}-8.5} & 59.7\rc{-13.3} & 80.3\rc{\hphantom{0}-6.0} & -21.5  \\
\rownumber Sentence order shuff.      & 72.2\rc{\hphantom{0}-6.8} & 56.1\rc{\hphantom{0}-4.0} & 53.7\rc{-15.6} & 90.3\rc{\hphantom{0}-1.3} & 80.7\rc{\hphantom{0}-1.5} & 50.3\rc{\hphantom{0}-4.5} & 82.5\rc{\hphantom{0}-6.0} & 75.6\rc{\hphantom{0}-3.0} & 66.8\rc{\hphantom{0}-2.9} & 85.4\rc{\hphantom{0}-0.0} & \hphantom{0}-4.6  \\
    \rowc  \rownumber Dummy numerics      & 75.9\rc{\hphantom{0}-1.9} & 57.8\rc{\hphantom{0}-1.0} & 60.0\rc{\hphantom{0}-5.6} & 89.5\rc{\hphantom{0}-2.2} & 78.7\rc{\hphantom{0}-3.9} & 49.7\rc{\hphantom{0}-5.7} & 85.0\rc{\hphantom{0}-3.2} & 76.2\rc{\hphantom{0}-2.3} & 67.8\rc{\hphantom{0}-1.5} & 85.3\rc{\hphantom{0}-0.1} & \hphantom{0}-2.8  \\
    \rownumber Logical words dropped       & 76.7\rc{\hphantom{0}-0.9} & 58.0\rc{\hphantom{0}-0.7} & 62.1\rc{\hphantom{0}-2.3} & 91.0\rc{\hphantom{0}-0.5} & 80.6\rc{\hphantom{0}-1.6} & 52.0\rc{\hphantom{0}-1.3} & 85.3\rc{\hphantom{0}-2.8} & 77.3\rc{\hphantom{0}-1.0} & 67.7\rc{\hphantom{0}-1.5} & 85.4\rc{\hphantom{-0}0.0} & \hphantom{0}-1.3  \\
   \rowc \rownumber Pronoun words dropped    & 76.5\rc{\hphantom{0}-1.2} & 57.0\rc{\hphantom{0}-2.5} & 63.4\rc{\hphantom{0}-0.3} & 91.2\rc{\hphantom{0}-0.2} & 81.8\rc{\hphantom{0}-0.2} & 52.0\rc{\hphantom{0}-1.3} & 86.6\rc{\hphantom{0}-1.4} & 77.4\rc{\hphantom{0}-0.8} & 68.3\rc{\hphantom{0}-0.7} & 84.8\rc{\hphantom{0}-0.8} & \hphantom{0}-0.9  \\
   \rownumber Causal words dropped         & 77.3\rc{\hphantom{0}-0.1} & 58.3\rc{\hphantom{0}-0.3} & 63.3\rc{\hphantom{0}-0.5} & 91.2\rc{\hphantom{0}-0.3} & 81.8\rc{\hphantom{0}-0.2} & 52.0\rc{\hphantom{0}-1.3} & 87.5\rc{\hphantom{0}-0.4} & 77.6\rc{\hphantom{0}-0.6} & 68.2\rc{\hphantom{0}-0.8} & 85.5\rc{\hphantom{-0}0.0} & \hphantom{0}-0.4  \\
    \bottomrule
  \end{tabular}
  \caption{
    The performances 
    (\%) of the baseline model with the ablation tests on the development set.
    Values in smaller font are changes (\%) relative to the original baseline performance, and the rightmost column (``Rel. avg.'') shows their averages.
  }
  \label{tlb:ablation}
\end{table*}

\subsection{Results of Reading and Reasoning Skills}
\label{sec:twolevels}

We report the results for the skills in Table \ref{tlb:ablation}.\footnote{In Appendix~E, we report dataset statistics and the average number of tokens dropped in each drop-based method.}
In the following, \% indicates a relative change from the original F1/accuracy unless specified otherwise.
In this section, we describe the notable findings for several skills.
The observations for all other skills are explained in Appendix~F.

\paragraph{$\boldsymbol{s_2}$ and $\boldsymbol{s_3}$: recognizing content words and function words.}
  On all datasets, the relative changes for $s_2$ were greater than those for $s_3$.
  However, it is remarkable that even with function words alone, the model could achieve 53.0\% and 17.4\% F1 on CoQA and SQuAD v1.1, respectively.\footnote{19.8\% of the questions in CoQA are yes/no questions.} 
  On ARC, RACE, and SWAG, the model showed more than 40\% accuracy ($>$25\% of random choice). 
%
  As for content words only, on all answer extraction datasets, the performance was greater than 78.7\% that of the original. 
  On all multiple-choice datasets, it was more than 90.2\%.
  These results imply that most of the questions already solved do not necessarily require grammatical and syntactic reasoning, in which function words are used.

\paragraph{$\boldsymbol{s_4}$: recognizing vocabulary beyond POS tags.}
Surprisingly, for SQuAD v1.1, the baseline model achieved 61.2\% F1.
It only uses 248 tokens as the vocabulary with the anonymization tags and no other actual tokens.
For the other answer extraction datasets, 
the largest drop (73.6\% relative) is by HotpotQA;
it has longer context documents than the other datasets, which seemingly makes its questions more difficult.
To verify the effect of its longer documents, we also evaluated the baseline model on HotpotQA without distracting paragraphs.
We found that the model's performance was 56.4\% F1 (the original performance was 76.3\% F1 and its relative drop was 26.1\%) which is much higher than that on the context with distracting paragraphs (16.8\% F1). This indicates that adding longer distracting documents contributes to encouraging machines to understand a given context beyond matching word patterns. 

On the other hand, the performance on the multiple choice datasets was significantly worse;
if multiple choices do not have sufficient word overlap with the given context, there is no way to infer the correct answer option.
Therefore, this result shows that multiple choice datasets might have a capacity for requiring more complex understanding beyond matching patterns between the question and the context than the answer extraction datasets. 

\paragraph{$\boldsymbol{s_6}$: recognizing the context word order (context words shuffle).}
We found that for the answer extraction datasets, the relative performance decreased by 55.6\% on average. 
A moderate number of questions are solvable even with the context words shuffled. 
We also found that, surprisingly, the average decrease was 21.3\% for the multiple choice datasets.
The drop on MCTest is more prominent than that on the others.
We posit that this is because its limited vocabulary makes questions more context dependent.
ARC, in contrast, uses factoid texts, and appears less context dependent.


\paragraph{$\boldsymbol{s_7}$: grasping sentence-level compositionality (sentence words shuffle).}
The performance with sentence words shuffled was greater than 60\% and 80\% those of the original dataset on the answer extraction and multiple-choice datasets, respectively.
This result means that most of the solved questions are solvable even with the sentence words shuffled.
However, we should not say that all questions must require this skill;
a question can require the performance of some complex reasoning (e.g., logical and multi-hop reasoning) and merely need to identify the sentence that gives the correct answer without precisely understanding that sentence.
Nevertheless, if the question is not intended to require such reasoning, we should care whether it can be solved with only a (sentence-level) bag of words.
In order to ensure that a model can understand the precise meaning of a described event, we may need to include questions to evaluate the grammatical and syntactic understanding into a dataset. 

\paragraph{$\boldsymbol{s_8}$: discourse relation understanding (sentence order shuffle).}
The smallest drop, excluding SWAG, which has one context sentence, was $-$1.3\%, on SQuAD v1.1.\footnote{\citet{min2018efficient} also reported that more than 90\% of questions on SQuAD v1.1 necessitate only a single sentence to answer them.}
Except for HotpotQA, the datasets show small drops (less than 10\%), which indicates that most solved questions do not require understanding of adjacent discourse relations and are solvable even if the sentences appear in an unnatural order. 

\paragraph{}
For SQuAD v2.0, we observed that the model recall increases for the no-answer questions.
Because F1 score is computed between the has- and no-answer question subsets, the scores tend to be higher than those for SQuAD v1.1.\footnote{See Appendix~G for detailed numbers.}

\subsection{Further Analyses} 
\label{sec:further}

To complement the observations in Section~\ref{sec:twolevels}, we performed further experiments as follows. 

\begin{table*}
  \footnotesize
  \setlength{\tabcolsep}{4pt}
  \def\arraystretch{1.0}
  \newcommand{\rc}[1]{\hspace{-0em}$_{\text{#1}}$}
  \newcommand{\linestack}[1]{\def\arraystretch{0.8}\begin{tabular}[c]{@{}c@{}} #1 \end{tabular}}
    \begin{tabular}{lcccccccccc|c} \toprule
      Ablation method \textbackslash ~Dataset & CoQA & DuoRC & \linestack{Hotpot- \\ QA} & \linestack{SQuAD \\ v1.1} & \linestack{SQuAD \\ v2.0} & ARC & MCTest & MultiRC & RACE & SWAG & \linestack{~Rel. \\ ~avg.} \\ \midrule
    \hphantom{00$'$.}  Original dataset    & 77.4\rc{\hphantom{-0}0.0} & 58.4\rc{\hphantom{-0}0.0} & 63.6\rc{\hphantom{-0}0.0} & 91.5\rc{\hphantom{-0}0.0} & 81.9\rc{\hphantom{-0}0.0} & 52.7\rc{\hphantom{-0}0.0} & 87.8\rc{\hphantom{-0}0.0} & 78.0\rc{\hphantom{-0}0.0} & 68.8\rc{\hphantom{-0}0.0} & 85.4\rc{\hphantom{-0}0.0} & \hphantom{-0}0.0  \\
    \midrule
    \rowc \hphantom{00$'$.} Drop all Q words & \hphantom{0}6.7\rc{-91.3} & 10.8\rc{-81.6} & 10.0\rc{-84.2} & 12.0\rc{-86.9} & 50.1\rc{-38.9} & 36.6\rc{-30.6} & 61.6\rc{-29.9} & 53.2\rc{-31.8} & 55.4\rc{-19.5} & 76.9\rc{-10.0} & -50.5 \\
    \hphantom{00$'$.} Drop all C words  & - & - & - & - & - & 40.3\rc{-23.6} & 32.5\rc{-63.0} & 61.7\rc{-20.9} & 41.0\rc{-40.4} & 71.7\rc{-16.0} & -32.8 \\
    \rowc \hphantom{00$'$.} Drop all C\&Q words     & - & - & - & - & - & 29.9\rc{-43.3} & 35.3\rc{-59.8} & 57.2\rc{-26.7} & 34.9\rc{-49.3} & 62.1\rc{-27.3} & -41.3 \\
    \midrule
    Trained \& evaluated on \\
    \rowc \hphantom{0}3$'$. Content words only   & 71.0\rc{\hphantom{0}-8.3} & 51.1\rc{-12.6} & 61.7\rc{\hphantom{0}-3.0} & 85.4\rc{\hphantom{0}-6.6} & 74.8\rc{\hphantom{0}-8.7} & 49.0\rc{\hphantom{0}-7.0} & 80.6\rc{\hphantom{0}-8.2} & 74.5\rc{\hphantom{0}-4.4} & 58.4\rc{-15.2} & 84.3\rc{\hphantom{0}-1.4} & \hphantom{0}-7.5  \\
    \hphantom{0}6$'$. Context word shuff.  & 52.9\rc{-31.7} & 40.2\rc{-31.2} & 46.1\rc{-27.4} & 68.0\rc{-25.7} & 80.6\rc{\hphantom{0}-1.7} & 46.6\rc{-11.5} & 55.3\rc{-37.0} & 70.1\rc{-10.2} & 54.7\rc{-20.5} & 83.6\rc{\hphantom{0}-2.1} & -19.9  \\
    \rowc \hphantom{0}7$'$. Sentence word shuff. & 68.3\rc{-11.8} & 47.7\rc{-18.4} & 66.8\rc{\hphantom{-0}5.0} & 82.4\rc{\hphantom{0}-9.9} & 80.3\rc{\hphantom{0}-2.0} & 47.7\rc{\hphantom{0}-9.6} & 75.0\rc{-14.6} & 73.6\rc{\hphantom{0}-5.6} & 59.2\rc{-14.0} & 84.0\rc{\hphantom{0}-1.6} & \hphantom{0}-8.2  \\
    \bottomrule    
  \end{tabular}
  \caption{
    Results of further analyses: the performance (\%) after dropping all question (``Q'') and/or context (``C'') words, and that of the baseline model both trained and evaluated on the modified inputs.
  }
  \label{tbl:further}
\end{table*}

\paragraph{The whole question and/or context ablation.}
To correctly interpret the result for $s_1$, we should know the performance on the \emph{empty questions}.
Likewise, for multiple-choice questions, the performance on the \emph{empty context} should be investigated to reveal biases contained in the answer options. 
Therefore, we report the baseline results on the whole question and/or context ablations.\footnote{This approach was already investigated by \citet{kaushik2018how}. However, there is no overlap in datasets between ours and those they analyzed other than SQuAD v1.1.}

Our results are reported in Table \ref{tbl:further}. 
Although the performance on SQuAD v2.0 was relatively high, we found that the model predicted \emph{no answer} for all of the questions (in this dataset, almost half of the questions are \emph{no answer}).
The other answer extraction datasets showed a relative drop of 80--90\%.
This result is not surprising since this setting forces the model to choose an answer span arbitrarily. 
On the multiple-choice datasets, on the other hand, the accuracies were higher than those of random choice (50\% for MultiRC and 25\% for the others), which implies that some bias exists in the context and/or the options.

\paragraph{Training and evaluating on the modified context.}
A question that was raised during the main analysis is what would happen if the model was trained on the modified input.
For example, given that the performance with the content words only is high, we would like to know the upper bound performance when the model is forced to ignore function words also during training.
Hence we trained the model with the ablations for the following skills: $s_3$ content words only; $s_6$ context word shuffle; and $s_7$ sentence word shuffle. 

The results are reported in the bottom rows of Table \ref{tbl:further}.
On almost all datasets, the baseline model trained on the ablation training set ($s_3'$, $s_6'$, and $s_7'$) displayed higher scores than that on the original training set ($s_3$, $s_6$, and $s_7$).
On CoQA, for instance, the relative change from the original score was only $-$8.3\% when the model was trained on $s_3$ content words only.
Although $s_3'$ and $s_7'$ with RACE were exceptions, their learning did not converge within the specified number of epochs.
We observed that for all datasets the relative upper bounds of performance were on average 92.5\%, 80.1\%, and 91.8\% for $s_3$, $s_6$, and $s_7$, respectively.
These results support our observations in Section~\ref{sec:twolevels}, that is, the questions allow solutions that do not necessarily require these skills, and thus fall short of testing precise NLU.
Even without tuning on the ablation training set, however, our methods can make an optimistic estimation of questions that are possibly dubious for evaluating intended skills.

\paragraph{Data leakage in BERT for SWAG.}
BERT's performance on SWAG is close to the performance by humans (88.0\%). 
However, the questions and corresponding options for SWAG are generated by a language model trained on the BookCorpus \cite{zhu2015aligning}, on which BERT's language model is also pretrained.
We therefore suspect that there is severe data leakage in BERT's language model as reported in \citet{zellers2019hellaswag}.
To confirm this issue, we trained a model without the context (i.e., the first given sentence). 
The accuracy on the development set, which was also without the context, was 74.9\% (a relative decrease of 12.2\%).
This result suggests that we need to pay more attention to the relations of corpora on which a model is trained and evaluated, but leave further analysis for future work.

\section{Qualitative Evaluation}
\label{sec:human}

In this section, we qualitatively investigate our ablation methods in terms of the human solvability of questions 
and the reconstructability of ablated features.  


We analyze questions of SQuAD v1.1 and RACE which cover both answering styles and are influential in the community.
We randomly sampled 20 questions from each dataset that are correctly solved (100\% F1 and accuracy) by the baseline model on the original datasets.
Our analysis covers four ablation methods ($\sigma_3$ content words only (involving $\sigma_{10,11,12}$), $\sigma_4$ vocabulary anonymization, $\sigma_6$ context word shuffle, and $\sigma_7$ sentence word shuffle) which provided specific insights in Section~\ref{sec:experiment}.

\subsection{Human Solvability after the Ablation}
\label{sec:solvable}

\begin{table}[t]
  \centering \footnotesize
    \setlength{\tabcolsep}{3pt}
  \newcommand{\linestack}[1]{\def\arraystretch{0.8}\begin{tabular}[c]{@{}c@{}} #1 \end{tabular}}
  \begin{tabular}{lcccc} \toprule
    Method \textbackslash~~Dataset & \multicolumn{2}{c}{SQuAD v1.1} & \multicolumn{2}{c}{RACE} \\ \midrule
    & Human & Baseline & Human & Baseline \\ \midrule
    3. Content words only      & \textbf{100.0} & 86.7 & \textbf{95.0} & 90.0 \\
    4. Vocab. anonymization    & 70.0  & \textbf{77.6} & 10.0 & \textbf{25.0} \\
    6. Context words shuff.    & 40.0  & \textbf{53.3} & 30.0 & \textbf{75.0} \\
    7. Sentence words shuff.   & 70.0  & \textbf{70.5} & 75.0 & \textbf{85.0} \\
    \bottomrule
  \end{tabular}
  \caption{Comparison of the human solvability and the baseline model's performance (\%) on questions that 
    are sampled from the ablation tests.}
  \label{tbl:human}
\end{table}

\paragraph{Motivation.} 
In Section~\ref{sec:experiment}, we observed that the baseline model exhibits remarkably high performance on some ablation tests.
To interpret this result, we investigate if a question is solvable by humans and the model.
Concretely, the question after the ablation can be (A) solvable by both humans and the model, (B) solvable by humans but unsolvable by the model, (C) unsolvable by humans but solvable by the model, or (D) unsolvable by both humans and the model.
For Case~A, the question is easy and does not require complex language understanding.
For Cases~B and C, the model may use unintended solutions because (B) it does not use the same solution as humans or (C) it \textit{cleverly} uses biases that humans cannot recognize.
For Case~D, the question may require the skill intended by the ablation method.
Although Cases~A to C are undesirable for evaluating the systems' skills,
it seems to be useful to distinguish them for further improvement of the dataset creation.
We therefore perform the annotation of questions with human solvability;
We define that a question is solvable if a reasonable rationale for answering the question can be found in the context.

\paragraph{Results.}
Table \ref{tbl:human} shows the human solvability along with the baseline model's performance on the sampled questions.
The model's performance is taken from the model trained on the original datasets except for the vocabulary anonymization method.
For the content words only on both datasets, the human solvability is higher than the baseline performance.
Although these gaps are not significant, we might be able to infer that the baseline model 
relies on content words more than humans (Case~B).
Given that the high performance of both humans and the baseline model, most of the questions fall into Case~A, i.e., they are easy and do not necessarily require complex reasoning involving the understanding of function words.

\begin{figure}[t]
  \footnotesize \centering
  \newcommand{\highlight}[1]{\textcolor{blue}{\textbf{#1}}}
  \newcommand{\highlightx}[1]{\textcolor{red}{#1}}
  \newcommand{\miniindent}{0.98\linewidth}
  \newenvironment{miniright}{
    \vspace{-0.3em}\begin{flushright}\begin{minipage}[t]{\miniindent}
  }{
    \end{minipage}\end{flushright}\vspace{-0.3em}
  }
  \newcommand{\empA}[1]{\textit{#1}}
  \newcommand{\empB}[1]{\textit{#1}}
  \centering \footnotesize
  \fbox{%
    \parbox{0.95\linewidth}{%
      \textbf{Original context}
      \begin{miniright}
        [...] By now you have probably heard about Chris Ulmer, the 26-year-old teacher in Jacksonville, Florida, who starts his special education class by calling up each student individually to give them much admiration and a high-five. 
        I couldn't help but be reminded of Syona's teacher and how she supports each kid in a very similar way. Ulmer recently shared a video of his teaching experience.
        All I could think was: how lucky these students are to have such inspirational teachers. [...]
      \end{miniright}
      \textbf{Context with shuffled context words}
      \begin{miniright}
        [...] their with and to kids combined , t always of ( has ) mean problems the palsy five cerebral that communication , her standard '' assess ( . teacher a a now gesture Florida admiration and , much calling Ulmer to individually ( of class his heard Jacksonville year special you up Chris greeting five ) congratulation by give education who , them or about probably the in by each - student high , old - - have starts 26 . I s she similar reminded be ' each t and in help ' kid teacher [...] 
      \end{miniright}
      \textbf{Question}
      \begin{miniright}
        What can we learn about Chris Ulmer?
      \end{miniright}
      \textbf{Options (the answer is in bold)}
      \begin{miniright}
       \textbf{(A) He praises his students one by one.} (B) He is Syona's favorite teacher. (C) He use videos to teach his students. (D) He asks his students to help each other.
      \end{miniright}
    }
  }
  \caption{Example of questions with shuffled context words from RACE. Although the question appears unsolvable for humans, the baseline model predicts the correct answer.}
  \label{fig:docshuffle}
\end{figure}

For the other three methods, the human solvability is lower than the baseline performance. 
This result indicates that the questions correctly solved only by the baseline model may contain unintended biases (Case~C).
For example, the gap in the context word shuffle of RACE is significant (30.0\% vs. 75.0\%).
Figure \ref{fig:docshuffle} shows a question that is unsolvable for humans but can be solved by the baseline model.
We conjecture that while humans cannot detect biases easily, the model can exploit biases contained in the answer options and their relations to the given context.




\subsection{Reconstructability of Ablated Features}
\label{sec:guessable}

\paragraph{Motivation.} 
We also seek to investigate the reconstructability of ablated features.
Even if a question falls under Case~A in the previous section, it might require the skill intended by the ablation;
If a reader is able to \textit{guess} the dropped information and uses it to solve the question, we cannot say that the question does not require the corresponding skill.
For example, even after dropping function words ($\sigma_3$), we might be able to guess which function word to fill in a cloze based on grammaticality and lexical knowledge.
Such \textit{reconstructable} features possibly exist for some ablation methods.
However, they are not critical if they are unnecessary for answering questions.
We can list the following cases: ablated features are ($\alpha$) unreconstructable and unnecessary, ($\beta$) unreconstructable and necessary, ($\gamma$) reconstructable and unnecessary, and ($\delta$) reconstructable and necessary.
To verify that ablation methods work, we need to confirm that there are few questions of Case~$\delta$.
The other cases are not critical to our observations in the main experiment.
We therefore perform the annotation with the following queries:
(i) \textit{are ablated features reconstructable?} and (ii) \textit{are reconstructable features really necessary for answering?}
When the answers for both queries are yes, a question is in Case~$\delta$.
In the annotation, we define that features in a question are reconstructable if the features existing around the rationale for answering the question are guessable.
We also require that these features are necessary to decide the answer if the correct answer becomes undecidable without them.


\paragraph{Results.}
For both datasets, the annotation shows that, not surprisingly, almost all features are unreconstructable in the shuffled sentence/context words and the vocabulary anonymization (except for one example in RACE).
When these questions are solvable / unsolvable by humans, we can say that features are unnecessary (Case~$\alpha$) / necessary (Case~$\beta$) for answering the questions.
In contrast, the annotators could guess function words for some questions even if these words are dropped (SQuAD: 55.0\% and RACE: 15.0\%).
The annotation of the necessity also shows that, however, reconstructable features (function words in this case) for all the questions are not necessary to answer them (i.e., Case~$\gamma$).
Therefore, we could not find any question in Case~$\delta$.
We report the annotation results in Appendix~H.
It is not easy for the annotator to completely ignore the information of reconstructed features.
We leave designing a solid, scalable annotation scheme for future work.

In summary, we found that almost all ablated features are unreconstructable.
Although for some questions ablated features are reconstructable for the content words only, these words are not necessarily required for answering the questions.
Overall, this result supports our observations in Section~\ref{sec:experiment}, i.e., questions already solved in existing datasets do not necessarily require complex language understanding.





\section{Discussion}
\label{sec:discussion}

In this section, we discuss two requirements for developing the MRC task as an NLU benchmark. 

\paragraph{The control of question solvability.}
Not to allow the model to focus on unintended objectives, we need to ensure that each question is unsolvable without its intended requisite skill.
Therefore, when benchmarking, we first need to identify necessary features whose presence determines the question's solvability. 
To identify them, we might need to perform ablation testing with humans. 
Further, we need to evaluate a model in both regular and ablation settings.
This is because a model may detect some biases that enable it to solve the question;
such biases can actually be false for humans and may be acquired by the model through overfitting to datasets.
Nonetheless, there is a case in which, even if we can identify necessary features, the model can have prior, true knowledge (e.g., world knowledge) of the correct answer. 
In this case, the model can answer the question without the context.
To avoid this circumvention, we may need to evaluate the model on fictional texts. 

\paragraph{Comprehensiveness of requisite skills.}
Another aspect of NLU benchmarking is the comprehensiveness of skills.
Our proposed approach can be expanded in two further directions: (i) inner-sentence and (ii) multiple-sentence levels.
For (i), we can focus on understanding of specific linguistic phenomena.
This includes logical and semantic understanding such as in FraCaS \cite{cooper1994fracas} and SuperGLUE \cite{wang2019superglue}.
To investigate particular syntactic phenomena, we might be able to use existing analysis methods \cite{marvin2018targeted}. 
For (ii), our skills can include complex/implicit reasoning, e.g., 
spatial reasoning 
\cite{weston2015bAbI} and lexically dependent causal reasoning \cite{sap2019atomic}. 
Although we do not need to include all of these skills in a single dataset, we need to consider the generalization of models across them. 

\section{Conclusion}
  Existing analysis work in MRC is largely concerned with evaluating the capabilities of \textit{systems}.
  By contrast, in this work, we proposed an analysis methodology for the benchmarking capacity of \textit{datasets}.
  Our methodology consists of input-ablation tests, in which each ablation method is associated with a skill requisite for MRC.
  We exemplified 12 skills and analyzed 10 datasets.
  The experimental results suggest that for benchmarking sophisticated NLU, datasets should be more carefully designed to ensure that questions correctly evaluate the intended skills.
  In future work, we will develop a skill-oriented method for crowdsourcing questions.

\section*{Acknowledgments}

We would like to thank Max Bartolo, Pasquale Minervini, and the anonymous reviewers for their insightful comments.
This work was supported by JSPS KAKENHI Grant Numbers 18H03297 and 18J12960 and JST ACT-X Grant Number JPMJAX190G.

\fontsize{9.0pt}{10.0pt} \selectfont
\bibliography{aaai20}
\bibliographystyle{aaai}

\appendix

\section{Our Defined Requisite Skills}
\label{app:skills}

\paragraph{Reading skills.}
As $s_2$ and $s_3$, we propose limiting the information available in the context by dropping content and function words respectively, which is intended to ascertain the extent to which a question depends on the given word type (e.g., a preposition \emph{in} before a time-related expression for a \emph{when} question).
Skill $s_5$ provides a heuristic of the relative levels of \emph{attention} between a question and the context. 
Skill $s_6$ is used to ensure that a model can extract the information conditioned on the word order.

\paragraph{Reasoning skills.} 
%
Skill $s_8$ is for the understanding of discourse relations between adjacent sentences, which relies on information given by the sentence order in the context. 
When we shuffle the sentence order, various relations, such as causality and temporality, 
are expected to be broken.
%
Skills $s_9$ to $s_{12}$ are defined more specifically; we drop tokens that explicitly emphasize important roles in specific skills such as \emph{if} and \emph{not} in logical reasoning.

\section{Experimental Details}
\label{app:spec}

In this section, we provide details of the specifications used in our experiments.

\paragraph{Datasets.}
For CoQA, since this dataset allows for \emph{yes}/\emph{no}/\emph{unknown} questions, we appended these words to the end of the context.
These special words were not allowed to be dropped.
Additionally, we appended the previous question-answer pair prior to the current question so that the model can consider the history of the QA conversation.
To compute the performance on SQuAD v2.0, we used the best F1 value that was derived from the predictions with a no-answer threshold of $0.0$.
For DuoRC, we used the ParaRC dataset (the official preprocessed version provided by the authors).
When training a model on DuoRC and HotpotQA, we used the first answer span; i.e., the document spans that have no answer span were not used in training.
For MCTest and RACE, we computed accuracy by combining MC160 with MC500 and Middle with High, respectively.
For MultiRC, which is allowed to have multiple correct options for a question, we cast a pair consisting of a question and one option as a two-option multiple choice (i.e., whether its option is true or false) and computed the micro-averaged accuracy for the evaluation.
The SWAG dataset is a multiple-choice task of predicting which event is most likely to occur next to a given sentence and the subject (noun phrase) of a subsequent event.
We cast the first sentence as the context and the subject of the second sentence as the question.
To compute F1 scores for the answer extraction datasets, we used the official evaluation scripts provided for the answer extraction datasets.

\begin{table}
  \centering \footnotesize
    \begin{tabular}{lp{10em}} \toprule
    Anonymization tag & POS tag or tokens \\ \midrule
    @noun\{ID\} & NN, NNS, NNP, NNPS \\
    @verb\{ID\} & VB, VBD, VBG, VBN, VBP, VBZ \\
    @adj\{ID\} & JJ, JJR, JJS \\
    @adv\{ID\} & RB, RBR, RBS \\
    @number\{ID\} & CD \\
    @wh\{ID\} & WDT, WP, WP\$, WRB \\
    @prep\{ID\} & IN, TO \\
    @punct\{ID\} & (punctuation except for the period tokens below) \\
    @period\{ID\} & . ! ? \\
    \bottomrule
  \end{tabular}
  \caption{
    Examples of anonymization tags and corresponding POS tags (OntoNotes 5 version of Penn Treebank tag set).
    We use \textit{@noun}, \textit{@verb}, \textit{@adj}, \textit{@adv}, and \textit{@number} for content words.
  }
  \label{tbl:anon-tags}
\end{table}

\paragraph{Ablation methods.}
For $\sigma_4$ vocabulary anonymization, we used the tags as shown in Table \ref{tbl:anon-tags} and \textit{@other} tags for the other POS tags. 
For $\sigma_{10}$ logical words dropped, as logic-related terms, we used the following: \emph{all, any, each, every, few, if, more, most, no, nor, not, other, same, some,} and \emph{than}.
For $\sigma_{12}$ causal words dropped, as causality-related terms, we used the following: \emph{as, because, cause, since, therefore,} and \emph{why}.
For $\sigma_3'$ training with content words only, we dropped function words as well as punctuation marks so that the model would see only content words.


We show examples of questions for the ablation method $\sigma_{4}$ in Figure \ref{fig:contentfunction}. 

\begin{figure*}[tb]
  \newcommand{\highlight}[1]{\textcolor{blue}{\textbf{#1}}}
  \newcommand{\highlightx}[1]{\textcolor{red}{#1}}
  \newcommand{\miniindent}{0.95\linewidth}
  \newenvironment{miniright}{
    \vspace{-0.3em}\begin{flushright}\begin{minipage}[t]{\miniindent}
  }{
    \end{minipage}\end{flushright}\vspace{-0.3em}
  }
  \centering \footnotesize

  \fbox{%
    \parbox{0.95\linewidth}{%
      $\rhd$ RACE / high6527.txt / Question 1 \\
      \textbf{Context}
      \begin{miniright}
        Having a great collection of books at home doesn ' t really mean that you are a person who has a passion for literature and reading . It can be a family inheritance or it can be just to impress people around you by the fact that you are a person of culture . On the other hand , there are many persons who cannot afford to buy books , because some of them are quite expensive , but who usually go to libraries and spend hours reading something that interests them a lot , or just borrow books to home . From my point of view , literature is very important in our life . For example , reading is a means of gaining culture and enriching our knowledge in different areas . It can help us have a great imagination and it makes things easier when it comes to make compositions on different themes . [...] 
      \end{miniright}
      \textbf{Context with dropped function words}
      \begin{miniright}
        \mysout{Having} \mysout{a} great collection \mysout{of} books \mysout{at} home \mysout{doesn} ' \mysout{t} really mean \mysout{that} \mysout{you} \mysout{are} \mysout{a} person \mysout{who} \mysout{has} \mysout{a} passion \mysout{for} literature \mysout{and} reading . \mysout{It} \mysout{can} \mysout{be} \mysout{a} family inheritance \mysout{or} \mysout{it} \mysout{can} \mysout{be} \mysout{just} \mysout{to} impress people around \mysout{you} \mysout{by} \mysout{the} fact \mysout{that} \mysout{you} \mysout{are} \mysout{a} person \mysout{of} culture . \mysout{On} \mysout{the} \mysout{other} hand , \mysout{there} \mysout{are} many persons \mysout{who} cannot afford \mysout{to} buy books , \mysout{because} \mysout{some} \mysout{of} \mysout{them} \mysout{are} quite expensive , \mysout{but} \mysout{who} usually go \mysout{to} libraries \mysout{and} spend hours reading something \mysout{that} interests \mysout{them} \mysout{a} lot , \mysout{or} \mysout{just} borrow books \mysout{to} home . \mysout{From} \mysout{my} point \mysout{of} view , literature \mysout{is} \mysout{very} important \mysout{in} \mysout{our} life . \mysout{For} example , reading \mysout{is} \mysout{a} means \mysout{of} gaining culture \mysout{and} enriching \mysout{our} knowledge \mysout{in} different areas . \mysout{It} \mysout{can} help us \mysout{have} \mysout{a} great imagination \mysout{and} \mysout{it} makes things easier \mysout{when} \mysout{it} comes \mysout{to} make compositions \mysout{on} different themes . [...] 
      \end{miniright}

      \textbf{Question}
      \begin{miniright}
        People who are fond of literature are those that \underline{\hspace{3em}} .
      \end{miniright}

      \textbf{Options}
      \begin{miniright}
        (A) \textbf{have much interest in reading} (B) keep many books (C) go to libraries every day (D) buy expensive books in the bookstore
      \end{miniright}

      \textbf{Prediction before and after dropping function words}
      \begin{miniright}
        (A) $\rightarrow$ (A)
      \end{miniright}
    }
  }

  \fbox{%
    \parbox{0.95\linewidth}{%
      $\rhd$ CoQA / source: mctest, id: 3tdxmtx3cbu3qs5x4zz64vf5jpxi67, turn-id: 6 \\
      \textbf{Context}
      \begin{miniright}
        On a snowy winter morning , the brown-haired lady saw a squirrel that was hurt . It only had three legs , and it looked hungry . She put some corn out for the squirrel to eat , but other bully squirrels came , too . The brown-haired lady started giving the little squirrel peanuts to eat . She gave some to the bully squirrels , too , so they would leave the three-legged squirrel alone . The winter snow melted and then it was spring . the grass turned green and the air was warm . Now , when the little squirrel with three legs would come to see the brown-haired lady with the peanuts , it would take the peanuts and dig a little hole and hide the peanuts for later . The squirrel would hold the peanut in its mouth and dig and dig and dig , and then it would put the peanut in the hole and pat it down with its little front paw . Then it would run back over to the brown-haired lady and get some more peanuts to eat . unknown yes no \\
        Was he hungry ? yes
      \end{miniright}
      \textbf{Context with dropped function words}
      \begin{miniright}
        \mysout{On} \mysout{a} snowy winter morning , \mysout{the} brown-haired lady saw \mysout{a} squirrel \mysout{that} \mysout{was} hurt . \mysout{It} \mysout{only} \mysout{had} three legs , \mysout{and} \mysout{it} looked hungry . \mysout{She} put \mysout{some} corn \mysout{out} \mysout{for} \mysout{the} squirrel \mysout{to} eat , \mysout{but} \mysout{other} bully squirrels came , \mysout{too} . \mysout{The} brown-haired lady started giving \mysout{the} little squirrel peanuts \mysout{to} eat . \mysout{She} gave \mysout{some} \mysout{to} \mysout{the} bully squirrels , \mysout{too} , \mysout{so} \mysout{they} would leave \mysout{the} three-legged squirrel alone . \mysout{The} winter snow melted \mysout{and} \mysout{then} \mysout{it} \mysout{was} spring . \mysout{the} grass turned green \mysout{and} \mysout{the} air \mysout{was} warm . \mysout{Now} , \mysout{when} \mysout{the} little squirrel \mysout{with} three legs would come \mysout{to} see \mysout{the} brown-haired lady \mysout{with} \mysout{the} peanuts , \mysout{it} would take \mysout{the} peanuts \mysout{and} dig \mysout{a} little hole \mysout{and} hide \mysout{the} peanuts \mysout{for} later . \mysout{The} squirrel would hold \mysout{the} peanut \mysout{in} \mysout{its} mouth \mysout{and} dig \mysout{and} dig \mysout{and} dig , \mysout{and} \mysout{then} \mysout{it} would put \mysout{the} peanut \mysout{in} \mysout{the} hole \mysout{and} pat \mysout{it} \mysout{down} \mysout{with} \mysout{its} little front paw . \mysout{Then} \mysout{it} would run back \mysout{over} \mysout{to} \mysout{the} brown-haired lady \mysout{and} get \mysout{some} \mysout{more} peanuts \mysout{to} eat . unknown yes no \\
        \mysout{Was} \mysout{he} hungry ? yes
      \end{miniright}

      \textbf{Question}
      \begin{miniright}
        What did the lady put out for the squirrel ?
      \end{miniright}

      \textbf{Answer}
      \begin{miniright}
        corn
      \end{miniright}
      \textbf{Prediction before and after dropping function words}
      \begin{miniright}
         corn $\rightarrow$ peanuts 
      \end{miniright}
    }
  }

  \caption{
    Examples of questions after the application of our ablation method $\sigma_3$ (content words only).
    Blacked-out text denotes words that were dropped by the ablation function.
  }
  \label{fig:contentfunction}
\end{figure*}

\begin{table}[t]
  \centering \footnotesize
  \def\arraystretch{0.8}
  \begin{tabular}{lccccc} \toprule
    Dataset & $d$ & $b$ & $lr$ & $ep$ \\ \midrule
    CoQA & 512 & 24 & 3 $\times$ 10$^{-5}$ & 2  \\
    DuoRC & 512 & 24 & 3 $\times$ 10$^{-5}$ & 2 \\
    HotpotQA & 512 & 24 & 3 $\times$ 10$^{-5}$ & 2 \\
    SQuAD v1.1 & 384 & 24 & 3 $\times$ 10$^{-5}$ & 2 \\
    SQuAD v2.0 & 384 & 24 & 3 $\times$ 10$^{-5}$ & 2 \\
    ARC  & 384 & 24 & 1 $\times$ 10$^{-5}$ & 4  \\
    MCTest & 512 & 16 & 2 $\times$ 10$^{-6}$ & 4 \\
    MultiRC & 512 & 24 & 2 $\times$ 10$^{-5}$ & 4 \\
    RACE & 512 & 32 & 1 $\times$ 10$^{-5}$ & 4 \\
    SWAG & 128 & 32 & 1 $\times$ 10$^{-5}$ & 4 \\
    \bottomrule
  \end{tabular}
  \caption{
    Hyperparameters used in the experiments, where $d$ is the size of the token sequence fed into the model, $b$ is the training batch size, $lr$ is the learning rate, and $ep$ is the number of training epochs.
    We set the learning rate warmup in RACE to 0.05 and 0.1 for the other datasets.
    We used stride = 128 for documents longer than $d$ tokens.
  }
  \label{tbl:hyperparam}
\end{table}

\begin{table}[t!]
  \scriptsize \centering
  \setlength{\tabcolsep}{3.0pt}
  \newcommand{\rc}[1]{\hspace{-0em}$_{\text{#1}}$}
  \newcommand{\linestack}[1]{\def\arraystretch{0.8}\begin{tabular}[c]{@{}c@{}} #1 \end{tabular}}
    \begin{tabular}{lccccccccc} \toprule
    Ablation method & CoQA & DuoRC & HotpotQA & SQuAD1.1 & SQuAD2.0 \\ \midrule
    \rowc 6.  Context w. shuff. & 29.8 (0.3) & 25.4 (0.4) & 23.6 (0.3) & 35.9 (0.3) & 52.4 (0.2) \\
    7.  Sent. w. shuff. & 53.0 (0.2) & 35.9 (0.3) & 43.1 (0.3) & 62.1 (0.3) & 64.4 (0.3) \\
    \rowc 8. Sent. ord. shuff. & 72.2 (0.2) & 56.1 (0.4) & 53.7 (0.3) & 90.3 (0.1) & 80.7 (0.1) \\
    \bottomrule \\[-1.2em]
    \toprule
    Ablation method & ARC & MCTest & MultiRC & RACE & SWAG \\ \midrule
    \rowc 6.  Context w. shuff. & 47.4 (1.9) & 47.2 (1.3) & 64.3 (0.2) & 51.7 (0.4) & 78.6 (0.2)  \\
    7.  Sent. w. shuff. & 46.4 (2.0) & 70.6 (1.6) & 71.4 (0.3) & 59.7 (0.1) & 80.3 (0.1)  \\
    \rowc 8. Sent. ord. shuff. & 50.3 (0.9) & 82.5 (1.4) & 75.6 (0.4) & 66.8 (0.3) & 85.4 (0.0)  \\
    \bottomrule
    \end{tabular}
    \caption{
      Ablation results with variances in parentheses for shuffle-related skills ($s_6$, $s_7$, and $s_8$) for five different runs.
    }
    \label{tbl:variance}
\end{table}

\begin{table}[t]
  \scriptsize \centering
  \setlength{\tabcolsep}{1.6pt}
  \def\arraystretch{0.9}
  \newcommand{\rc}[1]{\hspace{-0em}$_{~\text{#1}}$}
  \newcommand{\linestack}[1]{\def\arraystretch{0.8}\begin{tabular}[c]{@{}c@{}} #1 \end{tabular}}
  \begin{tabular}{lcccccccccc} \toprule
    Dataset & CoQA & DuoRC & HotpotQA & SQuAD1.1 & SQuAD2.0 \\ \midrule
    Text genre & various & movie & \multicolumn{3}{|c}{Wikipedia} \\
    \midrule
    \rowc Avg. \# Q tokens          & \hphantom{00}6.6    & \hphantom{00}8.7    & \hphantom{00}18.0   & \hphantom{0}11.7  & \hphantom{0}11.4  \\
    Avg. \# C tokens          & 344.0  & 691.3  & 1206.5 & 147.6 & 151.6 \\
    \rowc Avg. \# sentences in C    & \hphantom{0}18.8   & \hphantom{0}25.3   & \hphantom{00}47.8   & \hphantom{00}5.7   & \hphantom{00}6.1   \\
    \midrule
    \multicolumn{5}{l}{Avg. \# dropped tokens} \\
\rowc \hphantom{0}1. Q interrogatives only      &      \hphantom{00}5.8\rc{100.0} & \hphantom{00}7.7\rc{100.0} & \hphantom{00}16.8\rc{100.0} & \hphantom{0}10.7\rc{100.0} & \hphantom{0}10.4\rc{100.0} \\
\hphantom{0}2. Function words only      &  151.0\rc{100.0} & 357.1\rc{100.0} & \hphantom{0}606.4\rc{100.0} & \hphantom{0}76.6\rc{100.0} & \hphantom{0}78.5\rc{100.0} \\
\rowc \hphantom{0}3. Content words only             &  131.6\rc{100.0} & 305.6\rc{100.0} & \hphantom{0}366.3\rc{100.0} & \hphantom{0}50.8\rc{100.0} & \hphantom{0}52.7\rc{100.0} \\
\hphantom{0}5. Most sim. sent. only     &  300.0\rc{\hphantom{0}99.8} & 623.6\rc{\hphantom{0}97.8} & 1139.0\rc{100.0} & 105.0\rc{\hphantom{0}97.8} & 109.8\rc{\hphantom{0}98.5} \\
\rowc \hphantom{0}9. Dummy numerics      &  \hphantom{00}6.3\rc{\hphantom{0}93.2} & \hphantom{00}5.4\rc{\hphantom{0}85.9} & \hphantom{00}60.7\rc{100.0} & \hphantom{00}5.7\rc{\hphantom{0}86.3} & \hphantom{00}5.3\rc{\hphantom{0}83.4} \\
    10. Logical words drop.         &  \hphantom{00}6.7\rc{100.0} & \hphantom{00}8.0\rc{\hphantom{0}91.1} & \hphantom{000}8.6\rc{100.0} & \hphantom{00}2.1\rc{\hphantom{0}75.7} & \hphantom{00}2.5\rc{\hphantom{0}78.8} \\ 
   \rowc 11. Pronoun words drop.    &  \hphantom{0}19.7\rc{\hphantom{0}98.6} & \hphantom{0}49.4\rc{\hphantom{0}99.4} & \hphantom{00}22.3\rc{\hphantom{0}99.9} & \hphantom{00}2.3\rc{\hphantom{0}72.6} & \hphantom{00}1.9\rc{\hphantom{0}70.8} \\
    12. Causal words drop.           &  \hphantom{00}2.4\rc{\hphantom{0}84.4} & \hphantom{00}5.5\rc{\hphantom{0}88.4} & \hphantom{000}9.8\rc{\hphantom{0}99.0} & \hphantom{00}1.4\rc{\hphantom{0}66.5} & \hphantom{00}1.5\rc{\hphantom{0}69.5} \\
    \bottomrule \\[-1em]
    \toprule
    Dataset & ARC & MCTest & MultiRC & RACE & SWAG \\ \midrule
    Text genre & science & story & story & various & video \\
    \midrule
    \rowc Avg. \# Q tokens          & \hphantom{0}25.5 & \hphantom{00}9.2 & \hphantom{0}17.6 & \hphantom{0}11.1 & \hphantom{00}3.0 \\
    Avg. \# C tokens          & 131.4 & 247.7 & 339.9 & 326.8 & \hphantom{0}13.3\\
    \rowc Avg. \# sentences in C    & \hphantom{00}8.6 & \hphantom{0}20.1 & \hphantom{0}15.9 & \hphantom{0}19.8 & \hphantom{00}1.0  \\
    \midrule
    \multicolumn{5}{l}{Avg. \# dropped tokens} \\
\rowc \hphantom{0}1. Q interrogatives only      & 24.4\rc{100.0} & \hphantom{00}8.1\rc{100.0} & \hphantom{0}16.5\rc{100.0} & \hphantom{0}10.6\rc{100.0} & 2.9\rc{100.0} \\
\hphantom{0}2. Function words only      & 67.8\rc{100.0} & 106.5\rc{100.0} & 168.7\rc{100.0} & 146.5\rc{100.0} & 6.4\rc{100.0} \\
\rowc \hphantom{0}3. Content words only             & 46.5\rc{100.0} & 106.7\rc{100.0} & 113.6\rc{100.0} & 132.6\rc{100.0} & 5.4\rc{\hphantom{0}99.8} \\
\hphantom{0}5. Most sim. sent. only     & 89.3\rc{\hphantom{0}98.3} & 217.6\rc{\hphantom{0}99.7} & 299.4\rc{100.0} & 288.0\rc{\hphantom{0}99.8} & 0.0\rc{\hphantom{00}0.1} \\
    \rowc  \hphantom{0}9. Dummy numerics      & \hphantom{0}2.2\rc{\hphantom{0}53.0} & \hphantom{00}1.5\rc{\hphantom{0}67.5} & \hphantom{0}20.1\rc{100.0} & \hphantom{00}6.2\rc{\hphantom{0}90.0} & 0.1\rc{\hphantom{00}8.0} \\
    10. Logical words drop.         &  \hphantom{0}2.8\rc{\hphantom{0}73.2} & \hphantom{00}4.6\rc{\hphantom{0}97.5} & \hphantom{00}4.7\rc{\hphantom{0}95.7} & \hphantom{00}7.6\rc{\hphantom{0}97.2} & 0.1\rc{\hphantom{00}7.3} \\
   \rowc 11. Pronoun words drop.    &  \hphantom{0}1.8\rc{\hphantom{0}65.8} & \hphantom{0}22.0\rc{100.0} & \hphantom{0}13.5\rc{\hphantom{0}99.2} & \hphantom{0}19.4\rc{\hphantom{0}99.1} & 0.8\rc{\hphantom{0}54.1} \\
     12. Causal words drop.           & \hphantom{0}1.3\rc{\hphantom{0}56.7} & \hphantom{00}1.2\rc{\hphantom{0}51.2} & \hphantom{00}2.2\rc{\hphantom{0}87.3} & \hphantom{00}2.3\rc{\hphantom{0}78.2} & 0.1\rc{\hphantom{00}9.2} \\
    \bottomrule
  \end{tabular}
  \caption{
    Statistics of the datasets examined and average numbers of tokens dropped by our ablation methods $\sigma_i$ ($i=1, 2, 3, 5, 9,..., 12$).
    The tokens are counted after tokenization of the punctuation.
    Values in smaller font denote the proportion (\%) of questions that contain dropped tokens.
  }
  \label{tbl:stats}
\end{table}

\section{Hyperparameters of the Baseline Model}
\label{app:hyperparam}

Hyperparameters used in the baseline model are shown in Table \ref{tbl:hyperparam}.

\section{Performance Variances in Shuffle Methods}
\label{app:variance}

We report the variance for shuffling methods $s_6$ context words shuffle, $s_7$ sentence words shuffle, and $s_8$ sentence order shuffle in Table \ref{tbl:variance}.

\section{Statistics of the Examined MRC Datasets}
\label{app:stats}

Table \ref{tbl:stats} shows the statistics for the examined MRC datasets.

\section{Full Observations of the Main Results}
\label{app:full}

In this appendix, we describe the results for the reading and reasoning skills not mentioned in Section 4.2. 

\paragraph{$\boldsymbol{s_1}$: recognizing question words.}
For the first four answer-extraction datasets, the performance decreased by more than 70\%.
For the multiple-choice datasets, the performance decreased by an average of 23.9\%.

\paragraph{$\boldsymbol{s_5}$: attending to the whole context other than similar sentences.}
Even with only the most similar sentences, the baseline models achieved a performance level greater than half their original performances in 8 out of 10 datasets. 
In contrast, HotpotQA showed the largest decrease in performance.
This result reflects the fact that this dataset contains questions requiring multi-hop reasoning across multiple sentences.
  
\paragraph{$\boldsymbol{s_9}$--$\boldsymbol{s_{12}}$: various types of reasoning.}
For these skills, we can see that the performance drops were small; given that the drop for $s_3$ recognizing content words alone was under 20\%, we can infer that specific types of reasoning might not be critical for answering the questions.
Some types of reasoning, however, might play an essential role for some datasets: $s_9$ numerical reasoning in HotpotQA (whose questions sometimes require answers with numbers) and $s_{11}$ pronoun coreference resolution in DuoRC (consisting of movie scripts).

\section{Detailed Results of SQuAD v2.0}
\label{app:squad2}

We report the ablation results for has-answer and no-answer questions in SQuAD v2.0 in Table \ref{tbl:squad2}.

\begin{table}[t!]
  \footnotesize \centering
  \newcounter{rowcounter22}
  \setlength{\tabcolsep}{3.5pt}
  \def\arraystretch{1}
  \newcommand{\rc}[1]{\hspace{-0em}$_{\text{#1}}$}
  \newcommand{\rownumber}{\stepcounter{rowcounter22}\ifnum\value{rowcounter22}<10 \hphantom{0}\fi\arabic{rowcounter22}. }
  \newcommand{\linestack}[1]{\def\arraystretch{0.8}\begin{tabular}[c]{@{}c@{}} #1 \end{tabular}}
  \begin{tabular}{lll|l} \toprule
    Ablation method \textbackslash Subset & \linestack{Has-ans \\ 5928} & \linestack{No-ans \\ 5945} & \linestack{Total \\ 11873} \\ \midrule
    \hphantom{00.} Original dataset  & 82.6\rc{  0.0} & 79.9\rc{  0.0} & 81.9\rc{  0.0} \\
\rowc \rownumber Interrogatives in Q  & \hphantom{0}8.6\rc{-89.6} & 47.3\rc{-40.8} & 50.1\rc{-38.9} \\   
\rownumber Function words only  & \hphantom{0}0.4\rc{-99.5} & 99.6\rc{ 24.7} & 50.1\rc{-38.8} \\
\rowc \rownumber Content words only         & 65.6\rc{-20.5} & 81.2\rc{  1.6} & 73.5\rc{-10.3} \\
\rownumber Vocab. anonymization       & 41.9\rc{-49.3} & 76.9\rc{ -3.8} & 59.4\rc{-27.5} \\
\rowc \rownumber Most sim. sent. only & 69.2\rc{-16.2} & 83.2\rc{  4.1} & 72.8\rc{-11.1} \\
\rownumber Context words shuff.  & \hphantom{0}9.1\rc{-89.0} & 95.5\rc{ 19.5} & 52.4\rc{-36.1} \\    
\rowc \rownumber Sentence words shuff.       & 38.8\rc{-53.0} & 90.2\rc{ 12.9} & 64.6\rc{-21.2} \\
\rownumber Sentence order shuff.& 78.4\rc{ -5.1} & 81.9\rc{  2.5} & 80.3\rc{ -2.0} \\
\rowc \rownumber Dummy numerics             & 74.7\rc{ -9.6} & 82.0\rc{  2.6} & 78.7\rc{ -3.9} \\
\rownumber Logical words dropped   & 80.4\rc{ -2.6} & 80.0\rc{  0.1} & 80.6\rc{ -1.6} \\
\rowc \rownumber Dummy pronoun res.         & 82.0\rc{ -0.7} & 80.6\rc{  0.8} & 81.8\rc{ -0.2} \\
\rownumber Causal words dropped    & 82.1\rc{ -0.5} & 79.9\rc{  0.0} & 81.8\rc{ -0.2} \\
\midrule
\rowc \hphantom{00.} All Q words dropped & 10.8\rc{-86.9} & 17.7\rc{-77.9} & 50.1\rc{-38.9} \\
\midrule
\hphantom{00.} Trained \& evaluated on \\
    \rowc \hphantom{0}4$'$. Content words only   & 75.6\rc{\hphantom{0}-8.5} & 72.9\rc{\hphantom{0}-8.8} & 74.8\rc{\hphantom{0}-8.7} \\
    \hphantom{0}6$'$. Context words shuff.  & 63.6\rc{-22.9}            & 97.1\rc{\hphantom{0}21.5} & 80.6\rc{\hphantom{0}-1.7} \\
    \rowc \hphantom{0}7$'$. Sentence words shuff. & 75.5\rc{\hphantom{0}-8.5} & 79.3\rc{\hphantom{0}-0.8} & 80.3\rc{\hphantom{0}-2.0} \\
    \bottomrule
  \end{tabular}
  \caption{
    Results on the dev set of SQuAD v2.0 for subsets with normal (Has-ans) and no-answer (No-ans) questions.
  }
  \label{tbl:squad2}
\end{table}

\section{The Annotation Results} 

\begin{table}[t!]
  \centering \footnotesize
    \setlength{\tabcolsep}{3.5pt}
  \newcommand{\linestack}[1]{\def\arraystretch{0.8}\begin{tabular}[c]{@{}c@{}} #1 \end{tabular}}
  \begin{tabular}{lcccc|cccc} \toprule
    Method \textbackslash~~Dataset & \multicolumn{4}{c|}{SQuAD v1.1} & \multicolumn{4}{c}{RACE} \\
    & $\alpha$ & $\beta$ & $\gamma$ & $\delta$ & $\alpha$ & $\beta$ & $\gamma$ & $\delta$ \\ \midrule 
    3. Content words only      & .45  & .00  & .55 & .00 & .80 & .05 & .15 & .00 \\
    4. Vocab. anonymization    & .70  & .30  & .00 & .00 & .10 & .90 & .00 & .00 \\
    6. Context words shuff.    & .40  & .60  & .00 & .00 & .30 & .70 & .00 & .00 \\
    7. Sentence words shuff.   & .70  & .30  & .00 & .00 & .70 & .25 & .05 & .00 \\
    \bottomrule
  \end{tabular}
  \caption{
    Frequency of questions for Cases~$\alpha$ to $\delta$ for SQuAD v1.1 and RACE.
    Ablated features are ($\alpha$) unreconstructable and unnecessary, ($\beta$) unreconstructable and necessary, ($\gamma$) reconstructable and unnecessary, and ($\delta$) reconstructable and necessary.
    Questions for Case~$\delta$ are problematic for interpreting our main observations.
    }
  \label{tbl:reconst}
\end{table}

Table \ref{tbl:reconst} shows the frequency of questions for Cases~$\alpha$ to $\delta$ for SQuAD v1.1 and RACE.
See Section 5.2 for details.

\end{document}